\documentclass{article}
\usepackage[preprint]{neurips_2025}  
\usepackage[table]{xcolor} 
\usepackage{colortbl}       
\usepackage{graphicx}    
\usepackage{caption}     
\usepackage{array}       
\usepackage{booktabs}    
\usepackage{xcolor}
\usepackage{placeins}        
\usepackage{amsmath}         
\usepackage{amsfonts}        
\usepackage{amssymb}         
\usepackage{multirow}        

\usepackage{enumitem}
\usepackage{wrapfig}

\newcommand{\vs}{vs.}

\title{
{HRGS: Hierarchical Gaussian Splatting for Memory-Efficient High-Resolution 3D Reconstruction}
}

\author{
  Changbai Li\thanks{Equal contribution.} \\
  School of Artificial Intelligence \\
  Beihang University \\
  \texttt{changboli@buaa.edu.cn}
  \And
  Haodong Zhu\footnotemark[1] \\
  School of Artificial Intelligence \\
  Beihang University \\
  \texttt{HaodongZhu@buaa.edu.cn}
  \And
  Hanlin Chen\thanks{Corresponding author.} \\
  School of Computing \\
  National University of Singapore \\
  \texttt{hanlin.chen@u.nus.edu}
  \And
  Juan Zhang\footnotemark[2] \\
  School of Artificial Intelligence \\
  Beihang University \\
  \texttt{zhang\_juan@buaa.edu.cn}
  \And
  Tongfei Chen \\
  School of Artificial Intelligence \\
  Beihang University \\
  \texttt{tfchen@buaa.edu.cn}
  \And
  Shuo Yang \\
  School of Automation Science and Electrical Engineering \\
  Beihang University \\
  \texttt{shuo1yang@buaa.edu.cn}
  \And
  Shuwei Shao \\
  School of Automation Science and Electrical Engineering \\
  Beihang University \\
  \texttt{swshao@buaa.edu.cn}
  \And
  Wenhao Dong \\
  School of Astronautics \\
  Beihang University \\
  \texttt{ZB2315207@buaa.edu.cn}
  \And
  Baochang Zhang \\
  School of Artificial Intelligence \\
  Beihang University \\
  \texttt{bczhang@buaa.edu.cn}
}

\begin{document}

\maketitle

\begin{figure*}[h]
    \centering

    \includegraphics[width=1.0\linewidth]{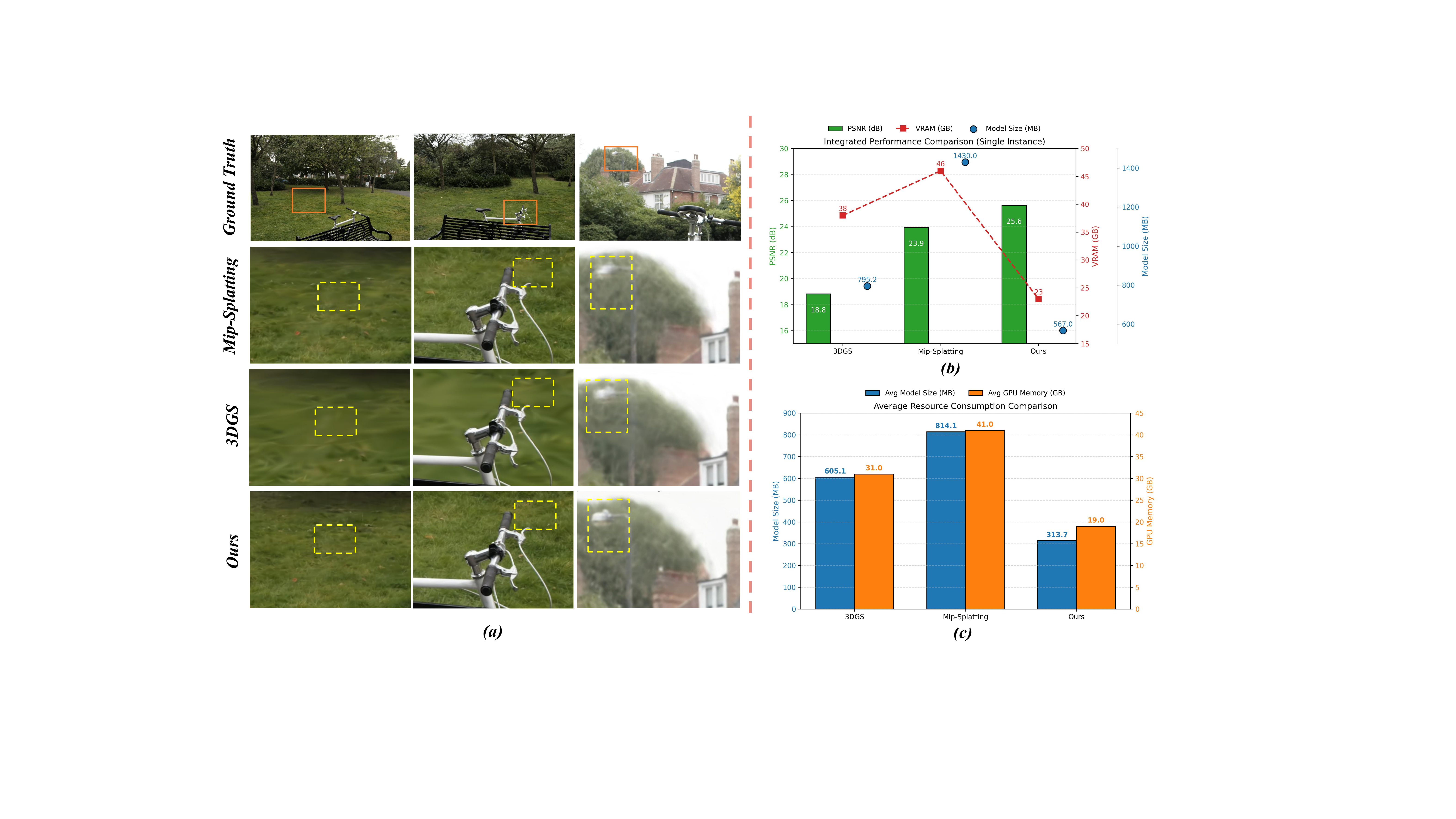}
    
    \caption{(a) High-resolution ($\sim$ 5K) renderings of the “bicycle” scene from the Mip-NeRF 360 dataset, with results from 3DGS, Mip-Splatting, and our method. Red dashed boxes highlight key details. (b) Performance on this scene: our method achieves the highest PSNR (25.6 dB) with significantly lower GPU memory (23 GB) and model size (567 MB) than 3DGS and Mip-Splatting. (c) Average resource usage across the full dataset shows our method maintains the smallest memory and model footprint.}
    %
    %
    \label{fig:top}
\end{figure*}


    

\begin{abstract}

3D Gaussian Splatting (3DGS) has achieved significant progress in real-time 3D scene reconstruction. However, its application in high-resolution reconstruction scenarios faces severe memory scalability bottlenecks. To address this issue, we propose Hierarchical  Gaussian Splatting (HRGS), a memory-efficient framework with hierarchical block-level optimization from coarse to fine. Specifically, we first derive a global, coarse Gaussian representation from low-resolution data; we then partition the scene into multiple blocks and refine each block using high-resolution data. Scene partitioning comprises two steps: Gaussian partitioning and training data partitioning. In Gaussian partitioning, we contract irregular scenes into a normalized, bounded cubic space and employ a uniform grid to evenly distribute computational tasks among blocks; in training data partitioning, we retain only those observations that lie within their corresponding blocks or make significant contributions to the rendering results. By guiding each block’s refinement with the global coarse Gaussian prior, we ensure alignment and seamless fusion of Gaussians across adjacent blocks. To reduce computational resource demands, we introduce an Importance-Driven Gaussian Pruning (IDGP) strategy: during each block’s refinement, we compute an importance score for every Gaussian primitive and remove those with minimal rendering contribution, thereby accelerating convergence and reducing redundant computation and memory overhead. To further enhance surface reconstruction quality, we also incorporate normal priors from a pretrained model. Finally, even under memory-constrained conditions, our method enables high-quality, high-resolution 3D scene reconstruction. Extensive experiments on three public benchmarks demonstrate that our approach achieves state-of-the-art performance in high-resolution novel view synthesis (NVS) and surface reconstruction tasks.
\end{abstract}

\section{Introduction}

3D scence reconstruction remains a longstanding challenge in computer vision and graphics. A significant advancement in this domain is the Neural Radiance Field (NeRF) \cite{nerf}, which effectively represents geometry and view-dependent appearance using multi-layer perceptrons (MLPs), demonstrating significant advancements in 3D reconstruction quality.

Recently, 3D Gaussian Splatting (3DGS)~\cite{3dgs} has gained considerable attention as a compelling alternative to MLP-based~\cite{nerf} and feature grid-based representations~\cite{tensorf,plenoxels,Neural,Instant_neural}. 3DGS stands out for its impressive results in 3D scene reconstruction and novel view synthesis while achieving real-time rendering at 1K resolutions. 
This efficiency and effectiveness, combined with the potential integration into the standard GPU rasterization pipeline, marks a significant step toward the practical adoption of 3D reconstruction methods.


    

In particular, 3DGS models complex scenes as a collection of 3D Gaussian distributions, which are projected onto screen space using splatting-based rasterization. The characteristics of each 3D Gaussian, including position, size, orientation, opacity, and color, are optimized using multi-view photometric loss. 
Although 3DGS has demonstrated impressive 3D reconstruction results, its application in high-resolution scenarios encounters critical memory scalability limitations. 
Specifically,when reconstructing outdoor scenes at ultra-high resolutions approaching 5K (e.g.4978\,$\times$\,3300 pixels) in standardized benchmark datasets like Mip-NeRF 360~\cite{Mip-NeRF360}, conventional 3DGS implementations demand excessive VRAM, exceeding the capacity of mainstream GPUs with limited memory, such as the NVIDIA A5000 (24GB VRAM). 
This computational bottleneck arises from the increasing resolution: higher resolutions demand more GPU memory, as illustrated in Fig.~\ref{fig:top}.
Such algorithmic behavior fundamentally conflicts with finite GPU memory resources, resulting in catastrophic memory overflow during optimization phases.

To overcome these critical memory constraints while preserving reconstruction fidelity for high-resolution scene reconstruction, we present Hierarchically  Gaussian Splatting (HRGS), a memory-efficient framework with hierarchical block optimization from coarse to fine. 
Specifically, we first obtain a coarse global Gaussian representation using low-resolution images. 
Subsequently, to minimize memory usage on a single GPU, we partition the scene into spatially adjacent blocks and parallelly refined. 
Each block is represented with fewer Gaussians and trained on reduced data, allowing further optimization with high-resolution images. The partitioning strategy operates at two levels: Gaussian primitives and training data.
To achieve a more balanced partition of Gaussians and avoid blocks with sparse Gaussians, we begin by contracting unbounded Gaussians. In detail, we define a bounded cubic region and use its boundary to normalize the Gaussian positions. Within this region, Gaussians are contracted via a linear mapping, while those outside undergo nonlinear contraction, yielding a more compact Gaussian representation. We then apply a uniform grid subdivision strategy to this contracted space, ensuring an even distribution of computational tasks.
During data partitioning for training, we compute the SSIM loss~\cite{mss} for each observation by comparing two renderings:One rendering is implemented with the complete global Gaussian representation, while the other is executed subsequent to the elimination of Gaussians within the target block. A more pronounced SSIM loss denotes that the observation exerts a more substantial contribution to the target block, so we set a threshold on SSIM loss and retain only observations whose values exceed it.
To mitigate artifacts at the block boundaries, we further include observations that fall within the region of the considered block.
Finally, to prevent overfitting, we employ a binary search algorithm during data partitioning to expand each block until the number of Gaussians it contains exceeds a specified threshold.
This innovative strategy effectively reduces interference from irrelevant data while improving fidelity with decreased memory usage, as demonstrated in Tab.~\ref{tab4}.

After partitioning the Gaussian primitives and data, we initialize each block in the original, uncontracted space using the coarse global Gaussian representation. To accelerate convergence and reduce computational overhead during block-level refinement with high-resolution data, we introduce an Importance-Driven Gaussian Pruning (IDGP) strategy. {Specifically, we evaluate the interaction between each Gaussian and the multi-view training rays within the corresponding block, and discard those with negligible rendering contributions. All blocks are then refined in parallel, and subsequently integrated into a unified, high-resolution global Gaussian representation.}
%

In addition to novel view synthesis, we evaluate our method on another key subtask of 3D reconstruction: 3D surface reconstruction. To further enhance the quality of the reconstructed surfaces, we incorporate the View-Consistent Depth-Normal Regularizer~\cite{vcR}, which is applied both during the initialization of the coarse global Gaussian representation and throughout the subsequent block-level refinement.
%
%




Finally, our method enables high-quality and high-resolution scene reconstruction even under constrained memory capacities (e.g., NVIDIA A5000 with 24GB VRAM).
We validate our method on two sub-tasks of 3D reconstruction: high-resolution NVS and surface reconstruction, and demonstrate that it delivers superior high-resolution reconstruction performance. In summary, the main contributions of this paper are:

\begin{itemize}[leftmargin=*]
    \item We propose HRGS, a memory-efficient coarse to fine framework that leverages low-resolution global Gaussians to guide high-resolution local Gaussians refinement, enabling high-resolution scene reconstruction with limited GPU memory.
    
    \item A novel partitioning strategy for Gaussian primitives and data is introduced, optimizing memory usage, reducing irrelevant data interference, and enhancing reconstruction fidelity.
    
     \item We propose a novel dynamic pruning strategy, Importance-Driven Gaussian Pruning (IDGP), which evaluates the contribution of each Gaussian primitive during training and selectively removes those with low impact. This approach significantly improves training efficiency and optimizes memory utilization.
    %
    %

    \item Extensive experiments on three public datasets demonstrate that our approach achieves state-of-the-art performance in high-resolution rendering and surface reconstruction.
\end{itemize}


\section{Related Work}

\noindent\textbf{3D Reconstruction.} 
Recent 3D reconstruction research can be broadly categorized into traditional geometry-based and deep learning methods. The former relies on multi-view stereo (MVS)~\cite{YAN} and structure from motion (SfM)~\cite{sfm} to estimate scene depth and camera poses, producing point clouds and subsequent surface meshes. The latter integrates implicit functions (e.g., SDF, Occupancy)~\cite{Tri} with volumetric rendering for high-fidelity reconstruction, as exemplified by Neural Radiance Fields (NeRF)~\cite{nerf}. However, NeRF-based approaches often struggle with real-time performance in large-scale or dynamic scenarios. In contrast, 3D Gaussian Splatting~\cite{3dgs} encodes scenes as 3D Gaussians (with position, scale, and color), using differentiable point-based rendering to achieve fast training and inference while balancing accuracy and quality. Balancing high fidelity, scalability, and real-time capability remains a key challenge in 3D reconstruction.
Within the field of 3D reconstruction, there are primarily two main sub-tasks: novel view synthesis (NVS) and surface reconstruction. 

\vspace{0.5em}

\noindent\textbf{Novel View Synthesis.} 
Novel View Synthesis (NVS) aims to generate a target image from an arbitrary camera pose, given source images and their camera poses~\cite{light, lum}. NeRF~\cite{nerf} integrates implicit representations with volume rendering\cite{volume, eff}, demonstrating impressive results in view synthesis. However, dense point sampling remains a major bottleneck for rendering speed. To address this, various methods accelerate NeRF by replacing the original multi-layer perceptrons (MLPs)~\cite{learining, deep} with discretized representations, such as voxel grids~\cite{dir}, hash encodings~\cite{Instant}, or tensor radiation fields~\cite{tensorf}. Additionally, some approaches~\cite{bake, merf} distill pretrained NeRFs into sparse representations, enabling real-time rendering. Recent advancements in 3D Gaussian Splatting (3DGS) have significantly improved real-time rendering, demonstrating that continuous representations are not strictly necessary. However, directly optimizing and rendering at high resolutions drastically increase memory overhead, making it challenging to achieve real-time reconstruction of high-quality scenes on mainstream GPUs with limited memory (24GB). Our approach specifically addresses this challenge by reducing the computational cost of high-resolution processing while preserving reconstruction fidelity. 

\vspace{0.5em}

\textbf{Multi-View Surface Reconstruction.} 
In recent years, multi-view reconstruction methods have evolved from traditional geometric approaches to neural implicit representations. Traditional multi-view stereo methods~\cite{re_MVS1,re_MVS2,re_MVS3,re_MVS4,re_MVS5,re_MVS6} primarily rely on extracting dense depth maps~\cite{re_depth1,re_depth2} from multiple images, fusing them into point clouds~\cite{re_point,re_point2}, and generating scene models through triangulation or implicit surface fitting~\cite{re_fit}. Although these techniques have been widely adopted in both academia and industry, they are often susceptible to artifacts and loss of detail due to matching~\cite{re_matching1} errors, noise, and local optimum issues during reconstruction. 
Recent advances in neural implicit representations, such as NeRF~\cite{re_nerf} and its SDF-based variants~\cite{NeuS,MonoSDF}, have shifted the reconstruction paradigm from explicit depth estimation toward learning continuous volumetric or surface fields directly from images. These approaches inherently model geometry and appearance jointly, offering better robustness to occlusions and textureless regions.
While neural implicit methods offer high-quality reconstructions, they remain computationally intensive and face scalability challenges. As an alternative, 3D Gaussian Splatting (3DGS)~\cite{3dgs} adopts an explicit representation by projecting anisotropic Gaussians onto image space, enabling efficient and differentiable rasterization~\cite{yifan2019differentiable}. Despite its ability to support real-time rendering with high visual fidelity, 3DGS often suffers from insufficient geometric supervision, particularly in sparse-view or large-scale scenarios~\cite{neusg}. To address this limitation, recent methods such as VCR-GauS~\cite{vcr-gaus}, Vastgaussian~\cite{vast}, and SuGaR~\cite{SuGaR} introduce view-consistent depth and normal constraints~\cite{turkulainen2024dnsplatter,bae2024dsine,yin2019enforcing}, significantly enhancing both reconstruction accuracy and convergence stability. These developments position 3DGS as a promising solution for high-resolution and scalable surface reconstruction.


\begin{figure*}[t]
  \centering
  \centering \includegraphics[width=1.0\textwidth]{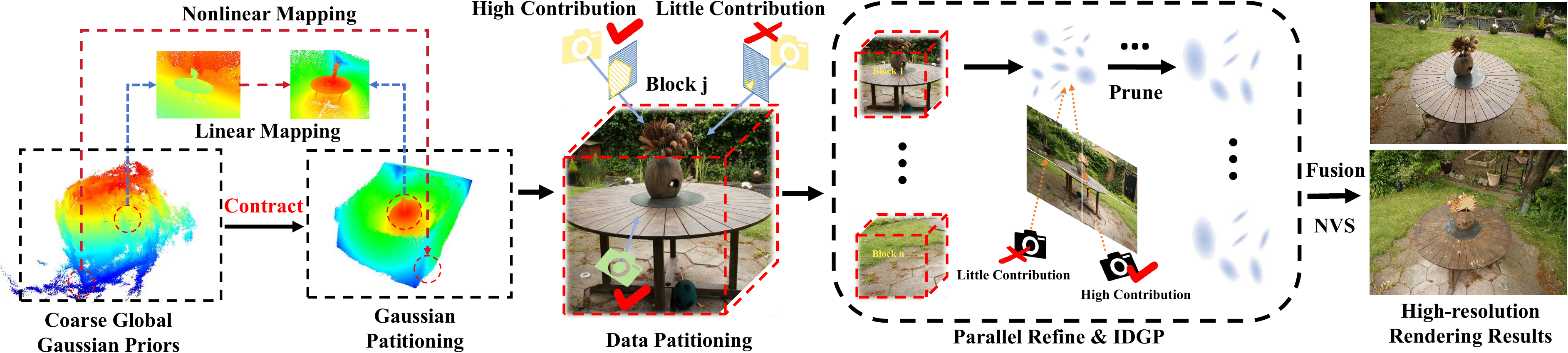}
   \caption{\textbf{Illustrative diagram of the hierarchical block optimization framework.} We first derive a global coarse Gaussian representation using low-resolution data, which is then contracted into a bounded cubic region. Subsequently, the contracted Gaussian primitives are partitioned into blocks, each paired with corresponding data. Leveraging the global coarse Gaussian as initialization, we parallelly refine each block in the original uncontracted space using high-resolution data. During this refinement process, an Importance-Driven Gaussian Pruning strategy is employed to compute the interaction between each Gaussian primitive and training view rays, removing low-contribution primitives to accelerate convergence and reduce redundancy. The optimized blocks are then concatenated to form the final global Gaussian representation, which is validated through novel view synthesis (NVS) and surface reconstruction tasks. }
   \label{fig:main}
\vspace{-0.2cm}
\end{figure*}

\section{Method}
Our proposed HRGS efficiently reconstructs high-resolution scenes. 
We first review 3DGS in Section~\ref{3.1}. 
Next, in Section~\ref{3.2}, we present the memory-efficient coarse-to-fine framework, detailing the partitioning of Gaussian primitives and data, along with the proposed Importance-Driven Gaussian Pruning (IDGP) strategy.
%
%
Finally, Section~\ref{3.3} describes the loss function employed in our approach.

\subsection{Preliminary}
\label{3.1}
We begin with a brief overview of 3D Gaussian Splatting (3DGS)~\cite{3dgs}. 
In the 3DGS framework, a scene is represented as a set of discrete 3D Gaussian primitives, denoted by \( G_K = \{ G_k \mid k = 1, \dots, K \} \), where \( K \) is the total number of Gaussians in the scene. 
Each Gaussian \( G_k \) is defined by a set of learnable parameters, including its 3D position \( \mathbf{p}_k \in \mathbb{R}^{3 \times 1} \), opacity \( \sigma_k \in [0, 1] \), and geometric properties, which typically consist of scaling and rotation parameters that define the Gaussian covariance matrix \( \Sigma_k \in \mathbb{R}^{3 \times 3} \). 
Furthermore, spherical harmonic (SH) features \( f_k \in \mathbb{R}^{3 \times 16} \) are used to encode view-dependent color information \( c_k \in \mathbb{R}^{3 \times 1} \), allowing for a realistic depiction of color variations as a function of the viewing angle.

For rendering purposes, the combined color and opacity contributions from multiple Gaussians at a given pixel are weighted according to their respective opacities. The color blending for overlapping Gaussians is computed as follows:
\begin{equation}
\hat{C} = \sum_{k \in M} c_k \alpha_k \prod_{j=1}^{k-1} \left( 1 - \alpha_j \right),
\label{eq1}
\end{equation}
where \( c_k \) and \( \alpha_k = \sigma_kG_k \) denote the color and density of the \( k \)-th Gaussian primitive, respectively.


\subsection{Hierarchical Block Optimization Framework}  
\label{3.2}

Traditional 3D Gaussian methods~\cite{3dgs,vcR} rely on global iterative optimization for scene reconstruction but struggle with memory inefficiency in high-resolution settings, such as the Mip-NeRF 360~\cite{mip} dataset. To address this, we propose a hierarchical optimization framework that balances coarse global representation and fine-grained local refinement, as shown in Fig~\ref{fig:main}. We first construct a low-resolution global Gaussian prior, guiding block-wise high-resolution optimization to enhance geometric detail while maintaining memory efficiency. This approach enables precise reconstruction under constrained memory conditions. The following subsections detail the coarse global Gaussian generation, Gaussian and data partitioning strategies, as well as refinement and post-processing procedures.

\noindent\textbf{Coarse Global Gaussian Representation.}
This stage establishes the foundation for subsequent Gaussian and data partitioning. Initially, we train the COLMAP~\cite{schoenberger2016mvs,schoenberger2016sfm} points using all observations at a low resolution for 30,000 iterations, generating a coarse representation of the global geometric structure. The resulting Gaussian primitives are represented as \( G_K = \{ G_k \mid k = 1, \ldots, K \} \), where \( K \) denotes the total number of Gaussians. In the following block-wise high-resolution refinement, this robust global geometric prior ensures that Gaussians are positioned accurately, thereby preventing drift and eliminating inter-block discontinuities, minimizing significant fusion artifacts.

\noindent\textbf{Primitives and Data Division.}
Directly applying uniform grid division in the original 3D space may lead to uneven Gaussian distribution in local regions (e.g. many nearly empty grid cells alongside overly dense ones). 
To address this imbalance, we define a bounded cubic region and contract all Gaussians within it. Within this region, the central one-third of the space is designated as the internal region, while the surrounding area is classified as the external region. The internal region is bounded by the minimum and maximum corner positions, $\mathbf{p}_{\text{min}}$ and $\mathbf{p}_{\text{max}}$, which define the limits of the central one-third of the entire region.
To standardize the representation of global Gaussians, we introduce a normalization step:
\(
\hat{\mathbf{p}}_k =  2\left( {\mathbf{p}_k - \mathbf{p}_{\text{min}}}\right)/\left({\mathbf{p}_{\text{max}} - \mathbf{p}_{\text{min}}} \right)-1.
\)
As a result, the coordinates of Gaussians located in the internal region are constrained within the range $[-1, 1]$. 
To achieve more effective contraction of the global Gaussians, we apply a linear mapping for the Gaussians in the internal region, while a nonlinear mapping is employed for the external region (as shown in Fig.~\ref{fig:main}). The final contraction step is performed using the function described in \cite{scanerf}:
\small{
    \begin{equation}
        \text{contract} (\hat{\mathbf{p}}_k) =
        \begin{cases}
        \hat{\mathbf{p}}_k, & \text{if } \|\hat{\mathbf{p}}_k\|_\infty \leq 1, \\
        \left( 2 - \frac{1}{\|\hat{\mathbf{p}}_k\|_\infty} \right) \frac{\hat{\mathbf{p}}_k}{\|\hat{\mathbf{p}}_k\|_\infty}, & \text{if } \|\hat{\mathbf{p}}_k\|_\infty > 1.
        \end{cases}
        \label{12}
    \end{equation}
}
The contracted space is then uniformly partitioned into $n$ blocks (the specific number of blocks used will be discussed further in Sec.~\ref{experiments}.), resulting in a more balanced Gaussian partitioning.
After partitioning the Gaussians, our objective is to ensure that each block is sufficiently trained. In other words, the training data assigned to each block should be highly relevant to the region it represents, focusing on refining the details within the block. To achieve this, we select observations and retain only those that contribute significantly to the visible content of the corresponding block in the rendering results. Since SSIM loss effectively captures structural differences and is somewhat robust to brightness variations~\cite{mss}, we use it as the foundation for our data partition strategy. Specifically, for the $j$-th block, the global Gaussians contained within it are represented as:
\(
G_{Kj} = \{ G_k \mid b_{j, \text{min}} \leq \text{contract}(\hat{\mathbf{p}}_k) < b_{j, \text{max}}, k = 1, \dots, K_j \},
\)
where $b_{j, \text{min}}$ and $b_{j, \text{max}}$ define the spatial bounds of the $j$-th block, and $K_j$ is the number of Gaussians contained within the block. 
The set of observations assigned to the \(j\)-th block is defined by the following formula:
\small{
\begin{equation}
\begin{aligned}
    \mathbf{P}^1_j = \text{Mask}\left(\mathcal{L}_{\text{SSIM}} \left( I_{G_K} (\boldsymbol{\tau}), I_{G_K \setminus G_{Kj}} (\boldsymbol{\tau}) \right ) > \epsilon \right )
    \odot \boldsymbol{\tau},
\end{aligned}
\label{13}
\end{equation}
}
where \( \text{Mask}(\cdot) \) generates an element-wise binary mask. Each element of the mask is set to 1 if it satisfies the condition inside the mask (i.e. the SSIM loss exceeds a threshold \( \epsilon \)), and 0 otherwise.
The term \( G_K \setminus G_{Kj} \) denotes the portion of the global set \( G_K \) excluding the block \( G_{Kj} \). \( \boldsymbol{\tau} \) is a matrix containing all camera poses, with each column \( \tau_i \) representing the \( i \)-th camera pose.
\( \odot \) is element-wise product operation.
And the resulting set \( \mathbf{P}^1_j \) represents the camera poses assigned to the \( j \)-th block. 

However, this strategy does not account for the projection of the considered block, which may lead to artifacts at the edges of the block. To address this issue, we further include poses that fall within the boundaries of the considered block:
\begin{equation}
    \mathbf{P}^2_j = \text{Mask}\left(b_{j, \text{min}} \leq \text{contract} (\hat{\mathbf{p}}_{\tau_i}) < b_{j, \text{max}} \right) \odot \boldsymbol{\tau}.
\label{14}
\end{equation}
where $\hat{\mathbf{p}}_{\tau_i}$ is the position under the world coordinate of pose $i$. The final assignment is:


\begin{equation}
    \mathbf{P}_j(\boldsymbol{\tau}, G_{Kj}) = \mathrm{Merge}\bigl(\mathbf{P}^1_j, \mathbf{P}^2_j\bigr),
\end{equation}
where $\mathrm{Merge}$ denotes the concatenate operator that removes any duplicate elements, ensuring only one copy of each element is retained.

To prevent overfitting, we employ a binary search method~\cite{binary} to incrementally expand $b_{j, \text{min}}$ and $b_{j, \text{max}}$ until $K_j$ exceeds a predefined threshold. 
Notably, this procedure is applied exclusively during the data partitioning phase for each block.

\noindent\textbf{Importance-Driven Gaussian Pruning (IDGP).}
After the Gaussian primitives and data division, we proceed to train each block in parallel in the original uncontracted space. Specifically, we first initialize each block using the coarse global Gaussian prior, and then fine‐tune each block using high‐resolution data as detailed in Sec.~\ref{3.3}. During block‐level optimization, we further accelerate convergence and reduce redundancy by applying a lightweight importance scoring and pruning strategy. Let $\mathcal{R}_b$ denote the set of all rays cast from the training views assigned to block $b$. For each Gaussian primitive $p_i$ in block $b$, we only consider its interactions with $\mathcal{R}_b$ and define the weighted hit count as
\small{
\begin{equation}
H_i \;=\;\sum_{r\in\mathcal{R}_b}\mathbf{1}(p_i \cap r)\,T_{i,r},
\text{where } T_{i,r} =\prod_{\substack{p_k \cap r\\\mathrm{depth}(p_k)<\mathrm{depth}(p_i)}}\bigl(1 - \alpha_k\bigr).
\end{equation}
}
Here, $\mathbf{1}(p_i \cap r)=1$ if and only if ray $r$ intersects $p_i$, and $T_{i,r}$ accumulates the transmission up to $p_i$ by all closer primitives $p_k$. We then compute the raw volume of $p_i$ as
\small{
$
v_i \;=\;\prod_{d=1}^3 s_{i,d},
$}
where each \(s_{i,d}\) is the scale factor of $p_i$ along the $d$-th spatial axis, and apply logarithmic compression
$
\widetilde{v}_i \;=\;\ln\bigl(1 + v_i\bigr).
$
Finally, we assign each primitive an importance score with its opacity $\alpha_i$:
$
S_i \;=\;\alpha_i\;\widetilde{v}_i\;H_i.
$
After evaluating $\{S_i\}$ for all primitives in the block, we sort them in descending order and remove the lowest 20\%. The remaining Gaussians, now both globally informed by the coarse prior and locally pruned of low‐impact points, continue through block‐level fine‐tuning. Finally, we select the fine‐tuned Gaussians within each block and, guided by the global geometric prior, concatenate the blocks to obtain the fine‐tuned global Gaussian. Through this process, the previously coarse global Gaussians are significantly enhanced in areas where they lacked detail.

\subsection{Loss Function}  
\label{3.3}
To optimize both the coarse and refined stages,the loss functions are defined as follows. First, we use the RGB loss $\mathcal{L}_{RGB}$ from 3DGS for the novel view synthesis task. To reconstruct scene surfaces, we enforce normal priors $\mathbf{N}$ predicted by a pretrained monocular deep neural network~\cite{BaeD24} to supervise the rendered normal map $\hat{\mathbf{N}}$ using $\mathbf{L1}$ and cosine losses:

\begin{equation}
    \mathcal{L}_n = \| \hat{\mathbf{N}} - \mathbf{N} \|_1 + (1 - \hat{\mathbf{N}} \cdot \mathbf{N}).
\label{eq2}
\end{equation}

Additionally, to effectively update Gaussian positions, we utilize the predicted normal $\mathbf{N}$ from the pretrained model to supervise the D-Normal $\overline{\mathbf{N}}_d$. The D-Normal is derived from the rendered depth by computing the cross-product of horizontal and vertical finite differences from neighboring points:

\begin{equation}
\overline{\mathbf{N}}_d = \frac{\nabla_v \mathbf{d} \times \nabla_h \mathbf{d}}{\left|\nabla_v \mathbf{d} \times \nabla_h \mathbf{d}\right|},
\label{eq5}
\end{equation}
where $\mathbf{d}$ represents the 3D coordinates of a pixel obtained via back-projection from the depth map. We then apply the D-Normal regularization from ~\cite{vcR}:

\begin{equation}
    \mathcal{L}_{dn} = w \cdot \left(\| \bar{\mathbf{N}}_d - \mathbf{N} \|_1 + (1 - \bar{\mathbf{N}}_d \cdot \mathbf{N})\right),
\label{eq8}
\end{equation}
where $w$ is a confidence term. The overall loss function integrates these components:

\begin{equation}
    \mathcal{L}_{total} = \mathcal{L}_{RGB} + \lambda_1 \mathcal{L}_s + \lambda_2 \mathcal{L}_n + \lambda_3 \mathcal{L}_{dn},
\label{eq9}
\end{equation}
where $\lambda_1$, $\lambda_2$, and $\lambda_3$ balance the individual terms. The term $\mathcal{L}_s$ is introduced to simplify depth computation, as described in~\cite{vcR}.

\begin{figure*}[t]
  \centering
  \centering \includegraphics[width=1.0\textwidth]{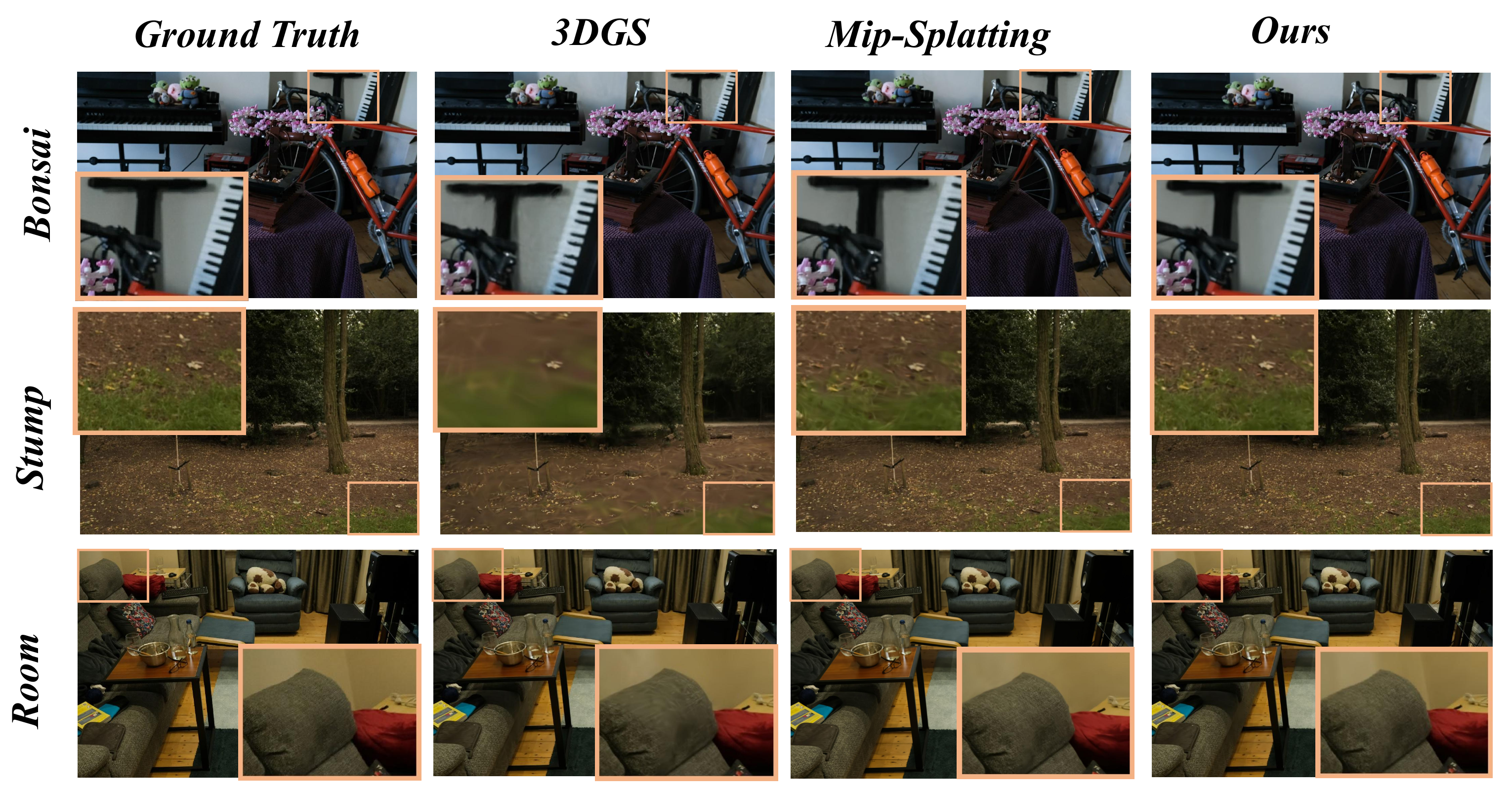}
  \caption{%
    \textbf{Qualitative Comparison on the Mip-NeRF 360 Dataset. }%
Three representative scenes demonstrate that our method more faithfully preserves fine-scale structures and achieves superior visual fidelity compared to 3DGS and Mip-Splatting.
  }
   \label{fig:4}
\end{figure*}

\definecolor{BurntOrange}{RGB}{204, 85, 0}

\begin{figure*}[t]
  \centering
  \centering \includegraphics[width=1.0\textwidth,height=0.2\textheight]{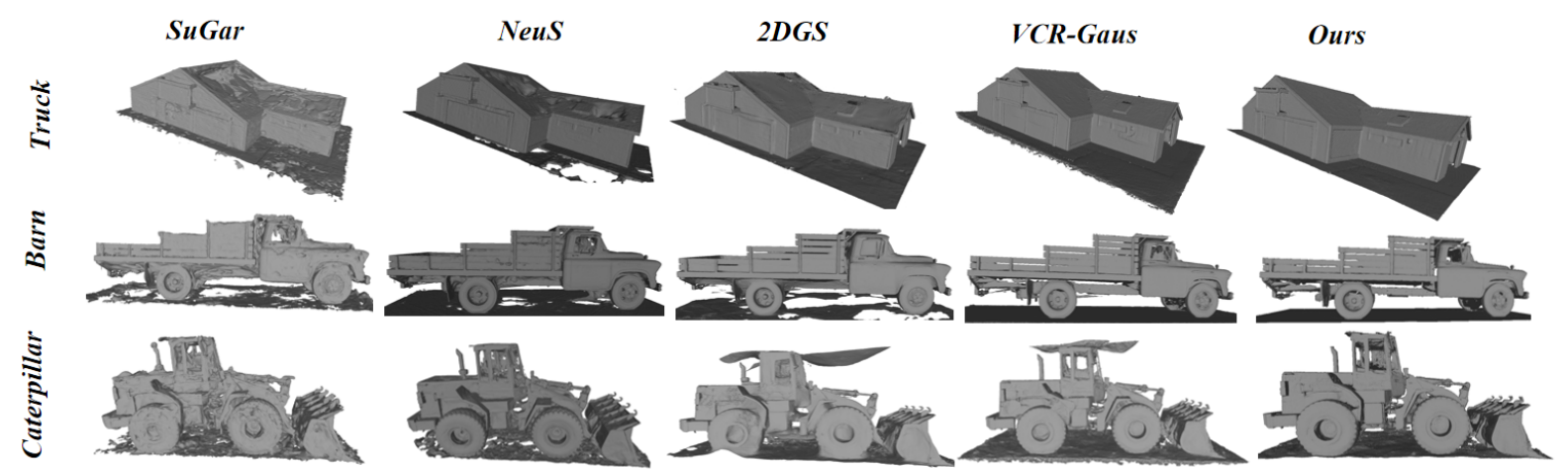}
   \caption{\textbf{Qualitative Comparison on TNT dataset.} Reconstructions from left to right—SuGar, NeuS, 2DGS, and VCR-Gaus—demonstrate that our method delivers more complete surface geometry, enhanced smoothness in planar regions, and superior preservation of fine structural details, thereby outperforming existing approaches in geometric fidelity. }
   \label{fig:3}
   \vspace{-0.2cm}
\end{figure*}

\section{Experiment}
\label{experiments}
\subsection{Experimental setups}
\noindent\textbf{Dataset and Metrics.}
To evaluate the effectiveness of our reconstruction method, we conduct experiments on two core tasks: novel view synthesis (NVS) and surface reconstruction, using multiple benchmark datasets. We first assess high-resolution NVS performance on Mip-NeRF360~\cite{mip}, followed by high-fidelity surface reconstruction on the Tanks and Temples (TNT)\cite{TNT} dataset. Additionally, we perform comparative experiments on the Replica\cite{replica} dataset to further validate our method. For a comprehensive evaluation, we employ standard metrics including SSIM, PSNR, LPIPS, and F1-score. Rendering efficiency is also assessed in terms of frames per second (FPS).
%
%

\noindent\textbf{Implementation Details.}
We begin by following the 3DGS~\cite{3dgs} pipeline, performing 30{,}000 iterations at a low resolution (0.3K) to obtain a coarse global Gaussian prior. During this stage, we introduce our Importance-Driven Gaussian Pruning (IDGP) strategy, which scores the rendering contribution of each Gaussian primitive and prunes those with the lowest impact. This step prevents irrelevant viewpoints from being assigned to training blocks in subsequent stages, reducing unnecessary computational overhead. The resulting coarse prior serves as initialization for the refinement phase.

In the contraction stage, we define the central one-third of the scene as the internal region and the remainder as the external region. The contracted Gaussians are then divided into four spatial sub-blocks. For data assignment, we use an SSIM threshold of $\epsilon = 0.1$. Each sub-block is further trained for 30{,}000 iterations. Specifically, we apply IDGP at the 10{,}000th, 15{,}000th, and 25{,}000th iterations to prune low-impact Gaussians based on their interaction contributions with training rays. This dynamic pruning accelerates convergence and reduces computational redundancy.

To facilitate surface reconstruction, we adopt the depth-normal regularization method described in Sec.~\ref{3.3}. Specifically, we use the pretrained DSINE~\cite{BaeD24} model for outdoor scenes and the pretrained GeoWizard~\cite{GeoWizard} for indoor scenes to predict normal maps. The hyperparameters $\lambda_1, \lambda_2,$ and $\lambda_3$ are set to 1, 0.01, and 0.015, respectively. After rendering the depth maps, we perform truncated signed distance function (TSDF) fusion and process the results using Open3D~\cite{Open3D}. 
Additional details are provided in the supplementary.

\begin{table}[t]
    \centering
    \caption{\textbf{Mip-NeRF 360 Full-Resolution Results.} The rendering quality comparison highlights the best and second-best results.}
    \label{tab:mipnerf_transposed}
    \setlength{\tabcolsep}{5pt}
    \begin{tabular}{@{}lccccccc@{}}
    \toprule
    & Mip-NeRF & Instant-NGP & zip-NeRF & 3DGS & 3DGS+EWA & Mip-Splatting & \textbf{Ours} \\
    \midrule
    PSNR $\uparrow$     & 24.10 & 24.27 & 20.27 & 19.59 & 22.81 & \textcolor{cyan}{26.22} & \textcolor{BurntOrange}{27.91} \\
    SSIM $\uparrow$     & 0.706 & 0.698 & 0.559 & 0.619 & 0.643 & \textcolor{cyan}{0.765} & \textcolor{BurntOrange}{0.863} \\
    LPIPS $\downarrow$  & 0.428 & 0.475 & 0.494 & 0.476 & 0.449 & \textcolor{cyan}{0.392} & \textcolor{BurntOrange}{0.342} \\
    \bottomrule
    \vspace{-0.4cm}
    \end{tabular}
    \label{tab1}
\end{table}

\begin{table*}[t]
    \centering
    \small
    \caption{\textbf{Quantitative Results on the Tanks and Temples Dataset \cite{TNT}.} The best results are highlighted in \textbf{\textcolor{BurntOrange}{orange}}, while the second-best results are marked in \textbf{\textcolor{cyan}{blue}}. }
    \setlength{\tabcolsep}{4pt}  
    \resizebox{0.88\textwidth}{!}{  
    \begin{tabular}{@{}lccccccc@{}}  
    \hline
    & \multicolumn{2}{c}{\textbf{NeuS-based}} & \multicolumn{5}{c}{\textbf{Gaussian-based}} \\  
    \cline{2-3} \cline{4-8}  
    \textbf{Scene} & NeuS & MonoSDF & SuGaR & 3DGS & 2DGS & VCR-GauS & Ours \\  
    \hline
    Barn & 0.29 & 0.49 & 0.14 & 0.13 & 0.36 & \textcolor{cyan}{0.62} & \textcolor{BurntOrange}{0.65} \\ 
    Caterpillar & \textcolor{cyan}{0.29} & \textcolor{BurntOrange}{0.31} & 0.16 & 0.08 & 0.23 & 0.26 & \textcolor{cyan}{0.29} \\ 
    Courthouse & 0.17 & 0.12 & 0.08 & 0.09 & 0.13 & \textcolor{cyan}{0.19} & \textcolor{BurntOrange}{0.23} \\ 
    Ignatius & \textcolor{BurntOrange}{0.83} & \textcolor{cyan}{0.78} & 0.33 & 0.04 & 0.44 & 0.61 & 0.64 \\ 
    Meetingroom & \textcolor{BurntOrange}{0.24} & \textcolor{cyan}{0.23} & 0.15 & 0.16 & 0.16 & 0.19 & \textcolor{BurntOrange}{0.24} \\ 
    Truck & 0.45 & 0.42 & 0.26 & 0.18 & 0.16 & \textcolor{cyan}{0.52} & \textcolor{BurntOrange}{0.61} \\ 
    \hline
    Mean & 0.38 & 0.39 & 0.19 & 0.09 & 0.30 & \textcolor{cyan}{0.40} & \textcolor{BurntOrange}{0.45} \\  
    Time & $>$24h & $>$24h & $>$1h & \textcolor{BurntOrange}{14.3m} & \textcolor{cyan}{34.2m} & 53m & 2h \\ 
    \hline
    FPS & \multicolumn{2}{c}{$<$10} & -- & \textcolor{BurntOrange}{159} & 68 & 145 & \textcolor{cyan}{146} \\  
    \hline
    \end{tabular}
    }
    \label{tab2}
\end{table*}


\noindent\textbf{Novel View Synthesis.}
As shown in Tab.~\ref{tab1}, we compare our method with several existing approaches, including mip-NeRF~\cite{Mip-NeRF360}, Instant-NGP~\cite{InstantNGP}, zip-NeRF~\cite{zip-nrf}, 3DGS~\cite{3dgs}, 3DGS+EWA~\cite{EWV}, and Mip-Splatting~\cite{mip360-splatting}. At high resolutions, our method significantly outperforms all state-of-the-art techniques.
As shown in Fig.~\ref{fig:4}, our method produces high-fidelity imagery devoid of fine-scale texture distortions.
While 3DGS~\cite{3dgs} introduces noticeable erosion artifacts due to dilation operations, Mip-Splatting~\cite{mip360-splatting} shows improved performance, yet still exhibits evident texture distortions. In contrast, our method avoids such issues, producing images that are both aesthetically pleasing and closely aligned with the ground truth, demonstrating the effectiveness of our hierarchicall refined strategy.

\begin{wraptable}{r}{0.42\textwidth}
    \small
    \vspace{-0.4cm}
    \centering
    \caption{\textbf{Comparsions on Replica~\cite{replica}.}} 
    \label{tab:3}
    \renewcommand{\arraystretch}{0.9}
    \setlength{\tabcolsep}{6pt}
    \begin{tabular}{llcc}
    \toprule
    \textbf{Type} & \textbf{Method} & \textbf{F1-score} & \textbf{Time} \\
    \midrule
    \multirow{2}{*}{Implicit}
      & NeuS      & 65.12 & \multirow{2}{*}{$>$10h} \\
      & MonoSDF   & \textbf{81.64} & \\
    \cmidrule{1-4}
    \multirow{4}{*}{Explicit}
      & 3DGS      & 50.79 & \multirow{4}{*}{$\leq$2h} \\
      & SuGar     & 63.20 & \\
      & 2DGS      & 64.36 & \\
      & Ours      & \textbf{74.87} & \\
    \bottomrule
    \end{tabular}
    \vspace{-0.4cm}
\end{wraptable}

\noindent\textbf{Surface Reconstruction.}
Our method not only provides high-quality novel views synthesis but also enables precise 3D surface reconstruction. 
As shown in Tab.~\ref{tab2}, our method outperforms both NeuS-based methods (e.g., NeuS~\cite{NeuS}, MonoSDF~\cite{MonoSDF}, and Geo-NeuS~\cite{Geo-Neus}) and Gaussian-based methods (e.g., 3DGS~\cite{3dgs}, SuGaR~\cite{SuGaR}, 2DGS~\cite{2Dgs}, and VCR-GauS~\cite{vcr-gaus}) on the TNT dataset. Compared to NeuS-based methods, our approach demonstrates significantly faster reconstruction speeds. While our method is slightly slower than some concurrent works, it achieves considerably better reconstruction quality, with a noticeable improvement over 2DGS (0.3 \vs 0.45). Furthermore, our method outperforms the recent state-of-the-art method in surface reconstruction, VCR-GauS, with a higher reconstruction quality (0.45 \vs 0.4). As shown in Fig.~\ref{fig:3}, our method is particularly effective at recovering finer geometric details. Additionally, our method has a substantial advantage in rendering speed, surpassing the rendering speed of 2DGS by more than double.
On the Replica dataset, as shown in Tab.~\ref{tab:3},
our method achieves comparable performance to MonoSDF~\cite{MonoSDF} while running significantly faster. 
Moreover, in comparison to explicit reconstruction approaches, 3DGS~\cite{3dgs}, SuGaR~\cite{SuGaR}, and 2DGS~\cite{2Dgs}, our method delivers substantially higher F1-scores.

\subsection{Ablation Studies}

To validate the effectiveness of individual components in our method, we conducted a series of ablation experiments on the ``Stump'' scene from the Mip-NeRF 360 dataset and the ``Ignatius'' scene from the TNT dataset. Specifically, we evaluated the impacts of the following components: hierarchical block optimization strategy, Importance-Driven Gaussian Pruning (IDGP), and data partitioning strategy.

\begin{table*}[h]
    \small
    \centering
    \caption{\textbf{Ablation on Data Division.} “SO Ass.” refers to SSIM-based assignment, while “BO Ass.” denotes boundary-based assignment. \textbf{Bold} indicates the best.}
    \label{tab4}
    \renewcommand{\arraystretch}{0.9}
    \setlength{\tabcolsep}{6pt}
    \begin{tabular}{lccccc}
    \toprule[1pt]
    \multirow{2}{*}{\textbf{Method}} & \multicolumn{5}{c}{\textbf{Settings}} \\
    \cmidrule{2-6}
     & baseline & w/o contraction & w/o SO Ass. & w/o BO Ass. & Full \\
    \midrule[0.8pt]
    PSNR $\uparrow$ & 21.87 & 22.73 & 24.29 & 22.96 & \textbf{25.24} \\
    F1 $\uparrow$    & 0.55  & 0.54  & 0.58  & 0.60  & \textbf{0.64} \\
    \bottomrule[1pt]
    \end{tabular}
    \label{tab4}
\end{table*}


\noindent\textbf{Ablation of the Data Division.}
As shown in Tab~\ref{tab4}, we analyzed the impact of the data partitioning strategy, using the original Gaussian global prior as the baseline. The results in the first and last rows of Table~\ref{tab5} demonstrate the effectiveness of our proposed method in improving performance (0.55 \vs 0.64). The second row in Table~\ref{tab5} further indicates that assigning relevant data in the contracted space is essential for enhancing reconstruction quality. The third row in Table~\ref{tab5} highlights the importance of starategy 1 (Eq.~\ref{13}) in data partitioning, and we also found that starategy 2 (Eq.~\ref{14}) plays a significant role in preventing artifacts at the edges of blocks.

\begin{wraptable}{r}{0.43\textwidth}
    \small
    \vspace{-0.4cm}
    \centering
    \caption{\textbf{Ablation on number of blocks.}} 
    \label{tab5_transposed}
    \renewcommand{\arraystretch}{0.9}
    \setlength{\tabcolsep}{6pt}
    \begin{tabular}{lcccc}
    \toprule[1pt]
    & \textbf{2} & \textbf{4} & \textbf{8} & \textbf{16} \\
    \midrule[0.8pt]
    \textbf{PSNR} $\uparrow$ & 24.86 & \textbf{25.24} & 23.93 & 22.56 \\
    \textbf{F1} $\uparrow$   & 0.62  & \textbf{0.64}  & 0.61  & 0.57  \\
    \bottomrule[1pt]
    \end{tabular}
    \label{tab5}
    \vspace{-0.3cm}
\end{wraptable}

\noindent\textbf{Ablation of the Number of Blocks.}
As shown in Tab.~\ref{tab5}, we investigate how the number of blocks affects reconstruction performance by splitting the coarse global Gaussian into 2, 4, 8, or 16 blocks. Our results indicate that too few blocks can cause conflicts between local and global optima, resulting in insufficient refinement of fine details, whereas too many blocks may lead to imbalanced data distribution and local overfitting. Consequently, we select four blocks for our experiments on the TNT dataset.

\begin{table}[h]
    \centering
    \small
    \caption{\textbf{Ablation Studies on the ``Stump'' Scene of the Mip-NeRF 360 Dataset\cite{Mip-NeRF360}.} }
    \label{tab:6}
    \begin{tabular}{@{}lccccc@{}}
    \toprule
    \textbf{Method} & \textbf{Model Size(MB)} & \textbf{GPU Memory(G)} & \textbf{PSNR} & \textbf{SSIM} & \textbf{LPIPS} \\
    \midrule
    Baseline           & 348.62        &   23  & 26.39 & 0.804 & 0.291 \\
    Baseline w/o IDPG &  601.04    & 28     & 26.41 & 0.791 & 0.288 \\
    \bottomrule
      \label{tab6}
    \end{tabular}
    \vspace{-0.2cm}
\end{table}

\noindent\textbf{Ablation of Importance-Driven Gaussian Pruning.}
As shown in Tab.~\ref{tab6}, to validate the effectiveness of the proposed Importance-Driven Gaussian Pruning (IDGP) strategy, we conducted an ablation study on the ``stump'' scene of the Mip-NeRF 360 dataset. 
Specifically, we compare the full method with the IDGP mechanism (Baseline) against a control variant in which IDGP is disabled (Baseline w/o IDGP).
%
%
As shown in Table.~\ref{tab6}, IDGP is able to selectively prune redundant Gaussians during training without degrading rendering quality, thereby achieving significant improvements in both model structure and computational efficiency.


\section{Conclusion}
In this work, we propose HRGS, a memory-efficient coarse-to-fine framework that uses low-resolution global Gaussians to guide the refinement of high-resolution local Gaussians, enabling high-resolution scene reconstruction under limited GPU memory. 
Our novel partitioning strategy for Gaussian primitives and data effectively facilitates block-wise optimization, significantly alleviating the high memory overhead typical of traditional 3DGS in high-resolution 3D reconstruction. Despite reduced memory requirements, our method achieves superior reconstruction quality and demonstrates state-of-the-art performance on two key sub-tasks of 3D reconstruction. 
As a result, this work establishes a new baseline for high-resolution 3D reconstruction, setting an important precedent for future research in the field.

\newpage
{\small
\bibliographystyle{ieee_fullname}
\bibliography{reference}
}

\newpage
\renewcommand\thefigure{\Alph{figure}} 
\setcounter{figure}{0}
\renewcommand\thetable{\Alph{table}} 
\setcounter{table}{0}
\renewcommand{\thesection}{\Alph{section}}
\setcounter{section}{0}

\setcounter{page}{1}

\definecolor{myorange}{RGB}{230,145,56}
\newcolumntype{Y}{>{\raggedleft\arraybackslash}X}
\newcommand{\ccell}[4]{#1/#2/#3/#4} 

\definecolor{myorange}{RGB}{230,145,56}
\section{Supplementary Material}
\label{sec:supplementary}







\begin{figure*}[ht!]
  \centering
  \centering \includegraphics[width=1.0\textwidth]{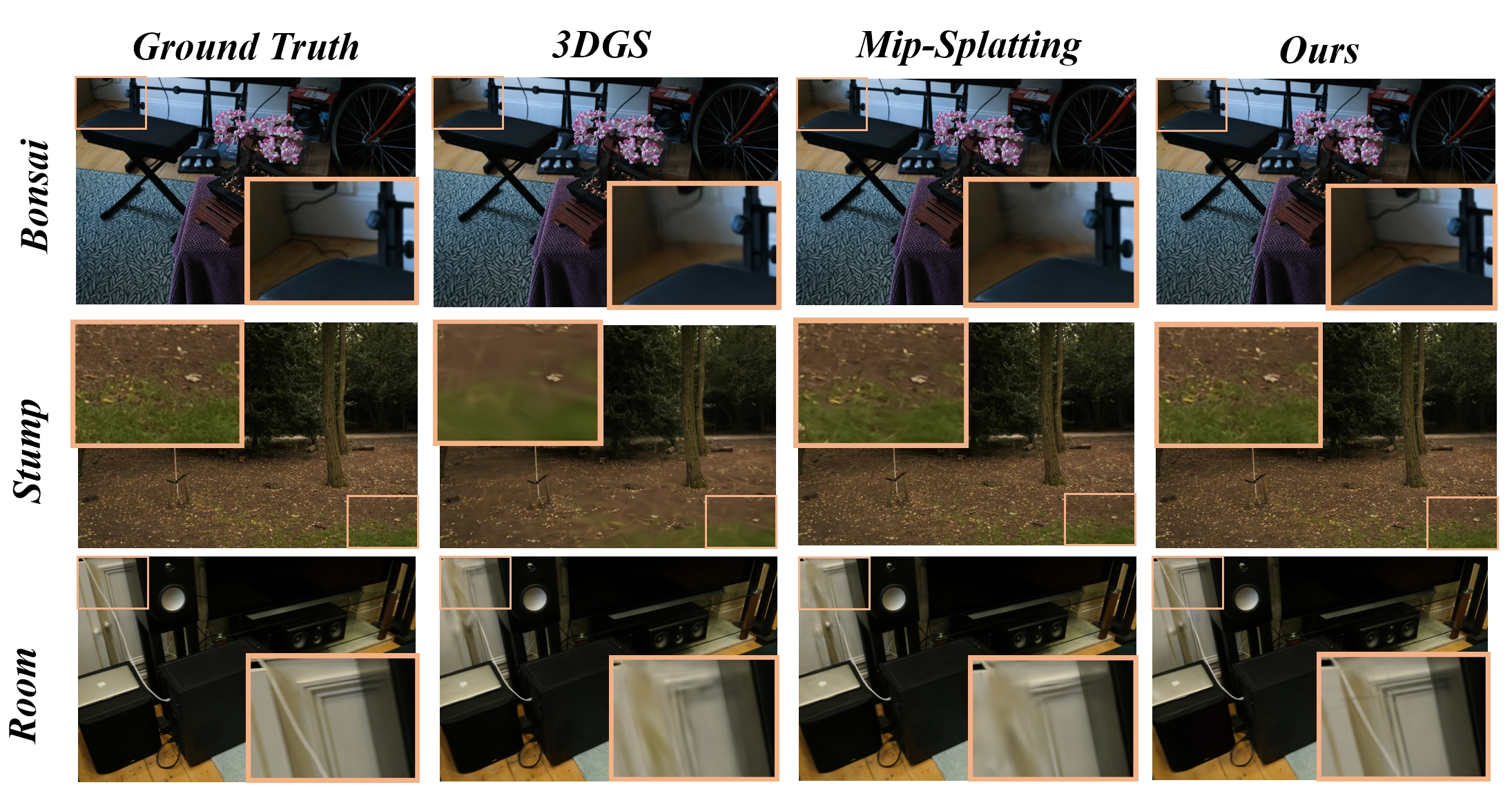}
  \caption{%
    \textbf{Qualitative comparison on the Mip-NeRF 360 Dataset.} Comparisons are conducted on the full-resolution dataset, showing that our method more faithfully preserves fine-scale structures and achieves superior visual fidelity compared to 3DGS and Mip-Splatting.}

   \label{fig:1}
\end{figure*}

\section{Implementation Details}

All experiments are conducted using machines equipped with NVIDIA A5000 GPUs, with PyTorch 2.0.1 and CUDA 11.7 as the software environment. For comparison methods that exceed the memory capacity of the A5000 setup, we employ NVIDIA A800 GPUs to ensure reliable execution and fair evaluation. Unless otherwise specified, we adopt the same hyperparameter settings as 3DGS~\cite{3dgs}. For outdoor scenes in the TNT dataset~\cite{TNT}, we incorporate decoupled appearance modeling~\cite{vast} to mitigate exposure-related artifacts. 
All models are trained and evaluated on the same data splits as used in 2DGS~\cite{2Dgs}, across the TNT~\cite{TNT} and Mip-NeRF360~\cite{Mip-NeRF360} benchmarks.

\section{Additional Ablation Study}
In this section, we further investigate the contributions of our proposed modules. We begin by analyzing the block-wise training strategy in Sec.\ref{sec:block_training}. Next, we examine the impact of the Data Division and IDGP strategies in Sec.\ref{sec:data_division} and Sec.\ref{sec:idgp}, respectively. While these ablations were previously conducted in the main paper, they were limited to a single scene. Here, we extend the evaluation to the full dataset, providing a more comprehensive and robust analysis. Finally, in Sec.\ref{sec:res_scale}, we assess the effectiveness of our method across various resolution scales.

\subsection{ Block-Wise Training for High-Resolution Reconstruction}
\label{sec:block_training}
We first validate our proposed block-wise training strategy in high-resolution scenarios by comparing it against the block partitioning scheme adopted in CityGS~\cite{Citygs}.
For fair comparison, we implemented the four-block division scheme following the methodology from the CityGS and conducted experiments on the full-resolution Mip-NeRF 360 dataset.

\begin{figure}[ht!]
  \centering
  \centering \includegraphics[width=1.0\textwidth,height=0.4\textheight]{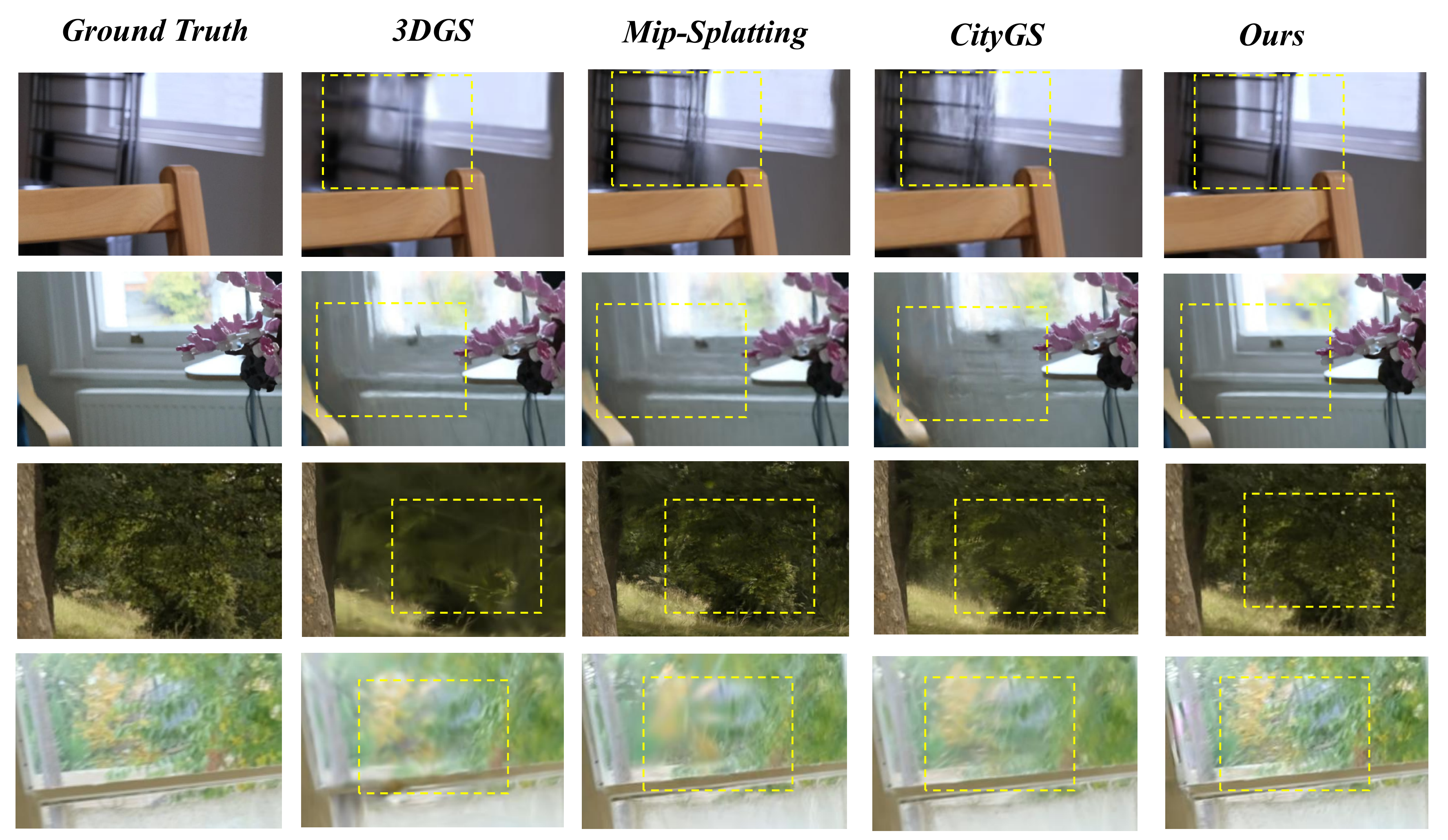}

   \caption{\textbf{Comparison of 3DGS, Mip-Splatting, CityGS and Ours at full resolution on the Mip-NeRF 360 dataset.}  }
   \label{fig:3}
\end{figure}

\begin{figure}[htbp]
  \centering
  \centering \includegraphics[width=0.7\textwidth,height=0.3\textheight]{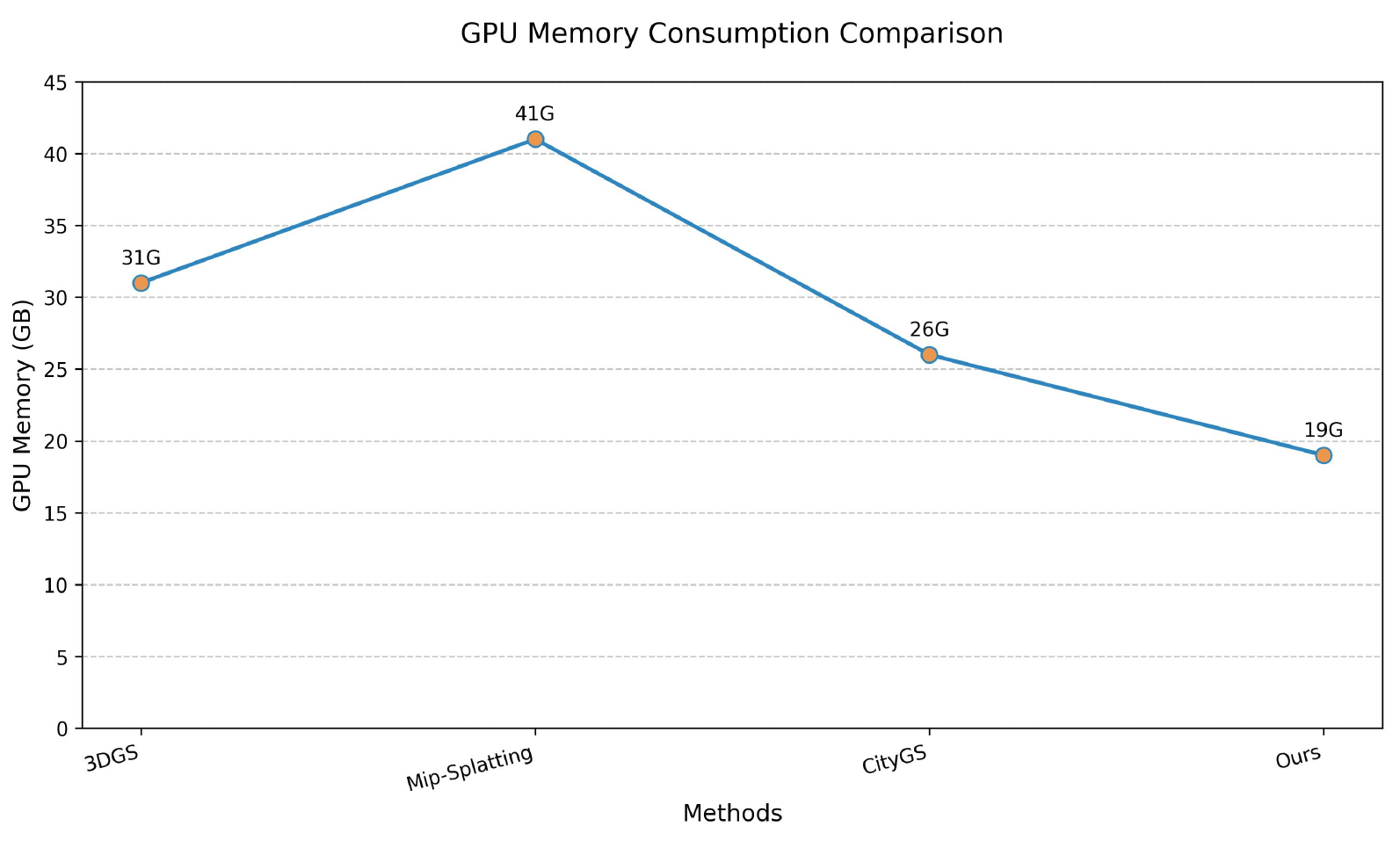}

   \caption{\textbf{Average GPU memory consumption of 3DGS, Mip-Splatting, CityGS, and our method on the Full-Resolution Mip-NeRF 360 dataset.}  }
   \label{fig:2}
\end{figure}

As shown in Tab.~\ref{tab:D}, our method achieves superior rendering quality compared to CityGS. As illustrated in Fig.~\ref{fig:3}, our approach demonstrates exceptional detail preservation in intricate geometric features and textural nuances. Additionally, Tab.~\ref{tab:cell} highlights the advantages of our block partitioning strategy over CityGS in enhancing model scalability. Across the Mip-NeRF 360 dataset, our method consistently reduces the total model size. For example, in the bicycle scene, the model size is 587.05 MB compared to the baseline's 1126.4 MB, with comparable improvements observed at the block level Tab.~\ref{tab:cell}. Furthermore, as shown in Fig.~\ref{fig:2}, our method substantially reduces GPU memory consumption compared to 3DGS, Mip-Splatting, and CityGS, underscoring its superior optimization of computational efficiency over CityGS despite employing similar block partitioning strategies.

\begin{table}[t]
    \centering
    \caption{\textbf{Mip-NeRF 360 Full-Resolution Results.} \textbf{Bold} indicates the best.}
    \vspace{2pt}
    \footnotesize
    \setlength{\tabcolsep}{3.5pt}
    \begin{tabular}{@{}lcccccccc@{}}
    \toprule
    & Mip-NeRF & Instant-NGP & zip-NeRF & 3DGS & 3DGS+EWA & CityGS & Mip-Splatting & \textbf{Ours} \\
    \midrule
    PSNR $\uparrow$     & 24.10 & 24.27 & 20.27 & 19.59 & 22.81 & 24.71 & {26.22} & {\textbf{27.91}} \\
    SSIM $\uparrow$     & 0.706 & 0.698 & 0.559 & 0.619 & 0.643 & 0.702 & {0.765} & {\textbf{0.863}} \\
    LPIPS $\downarrow$  & 0.428 & 0.475 & 0.494 & 0.476 & 0.449 & 0.401 & {0.392} & {\textbf{0.342}} \\
    \bottomrule
    \label{tab:D}
    \end{tabular}
    \vspace{-0.3cm}
\end{table}

\begin{table}[htbp]
\caption{\textbf{Comparison of model sizes (MB) across CityGS and Ours.} \textbf{Bold} indicates the best.}
\centering
\scriptsize
\begin{tabular}{c|ccccc|ccccc}
\hline
\multirow{2}{*}{Scene} & \multicolumn{5}{c|}{\textbf{CityGS (MB)}} & \multicolumn{5}{c}{\textbf{Ours (MB)}} \\
& Total & cell0 & cell1 & cell2 & cell3 & Total & cell0 & cell1 & cell2 & cell3 \\
\hline
bicycle & {1126.4} & 512.63 & 480.45 & 70.92 & 63.11 & \textbf{587.05} & 187.22 & 238.32 & 100.76 & 61.75 \\
bonsai  & {278.52} & 49.97  & 43.83  & 117.54 & 67.18 & \textbf{160.10} & 49.81  & 47.16  & 48.18  & 34.01 \\
counter & {215.85} & 21.38  & 22.26  & 78.73  & 93.49 & \textbf{144.58} & 23.06  & 36.63  & 36.36  & 45.54 \\
flowers & {738.18} & 269.72 & 276.41 & 96.18  & 93.87 & \textbf{399.71} & 163.86 & 133.96 & 65.55  & 63.33 \\
garden  & {1196.80} & 269.72 & 276.41 & 96.18  & 93.87 & \textbf{360.52} & 114.76 & 86.78  & 87.60  & 79.39 \\
kitchen & {324.08} & 30.26  & 26.46  & 157.15 & 110.21 & \textbf{170.94} & 38.41  & 35.78  & 76.55  & 47.52 \\
stump   & {873.41} & 399.37 & 302.45 & 171.59 & 191.51 & \textbf{368.62} & 103.86 & 103.32 &80.75  & 80.69 \\
treehill& {745.24} & 273.96 & 236.07 & 136.24 & 98.97 & \textbf{385.84} & 109.68 & 149.32 & 66.55  & 60.29 \\
room    & {301.02} & 33.58  & 94.33  & 74.08  & 106.03 & \textbf{221.97} & 316.31  & 33.84  & 75.64  & 76.19 \\
\hline
\end{tabular}
\label{tab:cell}
\end{table}

\subsection{ Ablation of the Data Division.}
\label{sec:data_division}

As shown in Table.~\ref{tab2}, we conducted chunk-based ablation experiments on the entire TNT dataset to evaluate the impact of data-partitioning strategies, using the original Gaussian global prior as the baseline. 
We systematically investigate the contribution of individual components, including contraction, SSIM-based assignment (SO Ass.) and boundary-based assignment (BO Ass.).

The results in the first and last columns of Tab.~\ref{tab2} demonstrate the effectiveness of our proposed method in improving performance (0.36 vs. 0.45). 
The second column in Tab.~\ref{tab2} further indicates that assigning relevant data in the contracted space is essential for enhancing reconstruction quality. 
The third column in Tab.~\ref{tab2} highlights the importance of SO Ass. in data partitioning, and we also found that BO Ass. plays a significant role in preventing artifacts at the edges of blocks.

\begin{table*}[h]
    \small
    \centering
    \caption{\textbf{Ablation on Data Division (performed on the full TNT dataset).} “SO Ass.” refers to SSIM-based assignment, while “BO Ass.” denotes boundary-based assignment. \textbf{Bold} indicates the best.}
    \label{tab4}
    \renewcommand{\arraystretch}{0.9}
    \setlength{\tabcolsep}{6pt}
    \begin{tabular}{lccccc}
    \toprule[1pt]
    \multirow{2}{*}{\textbf{Method}} & \multicolumn{5}{c}{\textbf{Settings}} \\
    \cmidrule{2-6}
     & baseline & w/o contraction & w/o SO Ass. & w/o BO Ass. & Full \\
    \midrule[0.8pt]
    PSNR $\uparrow$ & 22.19 & 23.63 & 24.89 & 23.96 & \textbf{25.02} \\
    F1 $\uparrow$    & 0.36  & 0.38  & 0.41  & 0.43  & \textbf{0.45} \\
    \bottomrule[1pt]
    \end{tabular}
    \label{tab2}
\end{table*}

\begin{figure*}[ht]
  \centering
  \centering \includegraphics[width=1.0\textwidth,height=0.55\textheight]{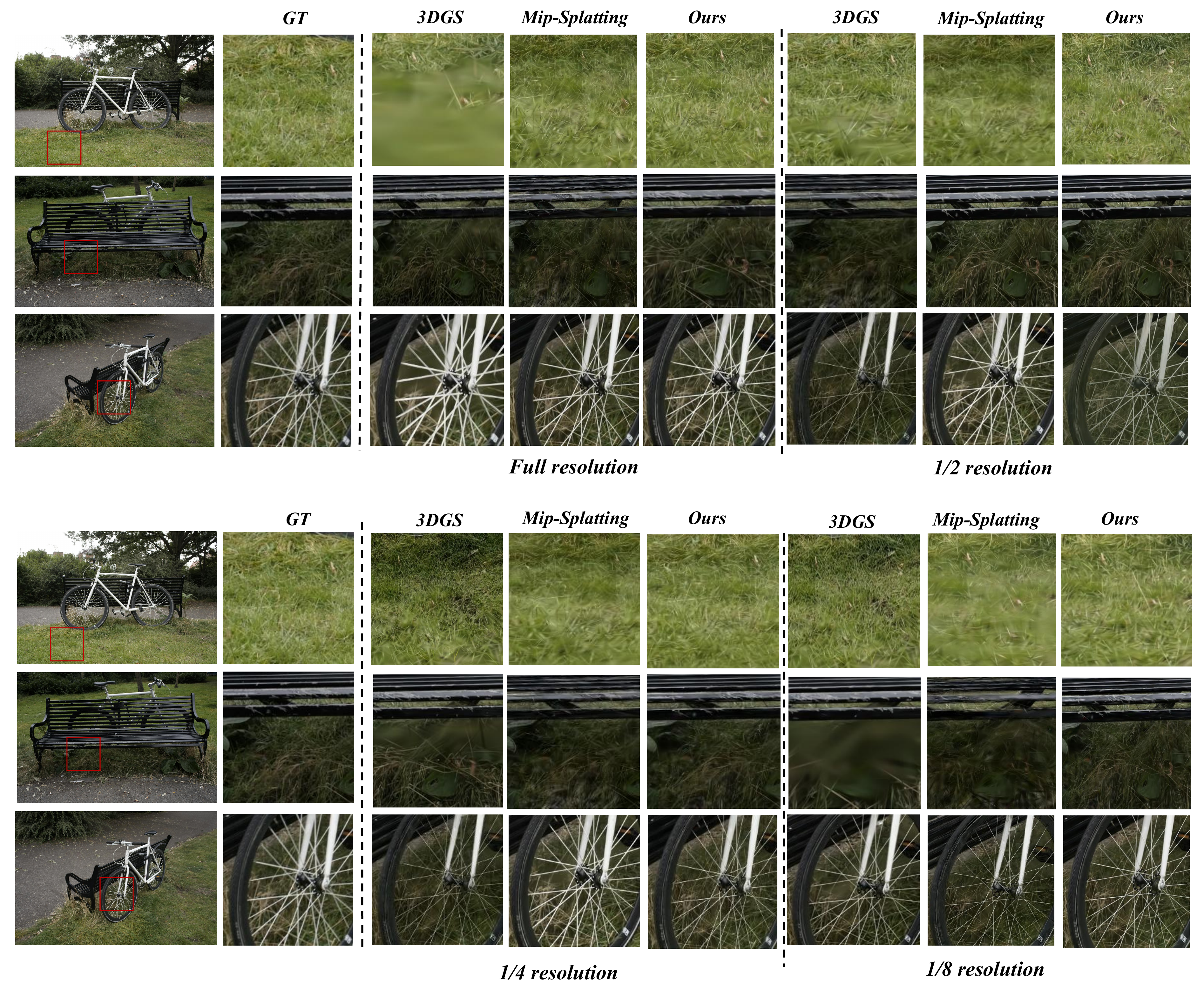}
   \caption{\textbf{Qualitative comparsion on Mip-NeRF 360 for different resolution scales.}  }
   \label{fig:4}
\end{figure*}

\subsection{Ablation of Importance-DrivenGaussian Pruning.}
\label{sec:idgp}

To evaluate the impact of Importance-Driven Gaussian Pruning (IDGP), we performed ablation studies on the full-resolution Mip-NeRF 360 dataset by comparing the proposed method with a variant that excludes IDGP. 
As shown in Tab.~\ref{tab6}, removing IDGP results in a substantial increase in model size from 313.72 MB to 621.04 MB, along with a 42\% rise in GPU memory usage (19 GB to 27 GB). Although PSNR slightly improves (28.02 vs. 27.91), the SSIM drops notably (0.821 vs. 0.863). These results underscore the effectiveness of IDGP in eliminating redundant Gaussians, significantly enhancing model efficiency while preserving high-quality reconstruction, thus highlighting its importance for memory-efficient 3D reconstruction in high-resolution settings.

\begin{table}[h]
    \centering
    \small
    \caption{{\textbf{Ablation study on IDPG across the entire full resolution Mip-NeRF 360 Dataset\cite{Mip-NeRF360}.}} }
    \label{tab:3}
    \begin{tabular}{@{}lccccc@{}}
    \toprule
    \textbf{Method} & \textbf{Model Size(MB)} & \textbf{GPU Memory(G)} & \textbf{PSNR} & \textbf{SSIM} & \textbf{LPIPS} \\
    \midrule
    w/o IDPG &  621.04    & 27     & 28.02 & 0.821 & 0.308 \\
    Ours (Full)           & 313.72        &   19  & 27.91 & 0.863 & 0.342 \\
    \bottomrule
      \label{tab6}
    \end{tabular}
    \vspace{-0.2cm}
\end{table}

\subsection{Robustness Validation Across Resolution Scales}
\label{sec:res_scale}

To thoroughly evaluate the robustness of our method across varying resolution scales, we performed detailed visual comparisons of rendering quality on the Mip-NeRF 360 dataset at multiple resolution levels.
As shown in Tab.~\ref{tab1}, we compare our method with several existing approaches, including mip-NeRF~\cite{Mip-NeRF360}, Instant-NGP~\cite{InstantNGP}, zip-NeRF~\cite{zip-nrf}, 3DGS~\cite{3dgs}, 3DGS+EWA~\cite{EWV}, and Mip-Splatting~\cite{mip360-splatting}. 
Our approach demonstrates comparable performance to these prior methods at one-eighth of the original resolution. 
Furthermore, at higher resolutions, our method significantly outperforms all state-of-the-art techniques.
As shown in Fig.~\ref{fig:4}, our method produces high-fidelity imagery devoid of fine-scale texture distortions.
While 3DGS~\cite{3dgs} introduces noticeable erosion artifacts due to dilation operations, Mip-Splatting~\cite{mip360-splatting} shows improved performance, yet still exhibits evident texture distortions. In contrast, our method avoids such issues, producing images that are both aesthetically pleasing and closely aligned with the ground truth, demonstrating the effectiveness of our hierarchicall refined strategy.

\begin{table*}[!ht]
\caption{\textbf{Quantitative results on Mip-NeRF 360 (Downscaled Resolutions).} The best results are highlighted in \textbf{\textcolor{BurntOrange}{orange}}, while the second-best results are marked in \textbf{\textcolor{cyan}{blue}}. Our method demonstrates consistent superiority across different resolution scales.}  
\vspace{-0.3cm}
\label{tab1}
\centering  
\vspace{0.2cm}  
\scalebox{0.85}{
\begin{tabular}{lccc|ccc|ccc}
\hline
\textbf{} & \multicolumn{3}{c|}{\textbf{1/2x Res.}} & \multicolumn{3}{c|}{\textbf{1/4x Res.}} & \multicolumn{3}{c}{\textbf{1/8x Res.}}  \\ 
\textbf{} & PSNR $\uparrow$ & SSIM $\uparrow$ & LPIPS $\downarrow$ & PSNR $\uparrow$ & SSIM $\uparrow$ & LPIPS $\downarrow$ & PSNR $\uparrow$ & SSIM $\uparrow$ & LPIPS $\downarrow$\\ \hline
mip-NeRF & 24.16 & 0.670 & 0.370 & 25.18 & 0.727 & 0.260 & 29.26 & 0.860 & 0.122  \\ 
Instant-NGP & 24.27 & 0.626 & 0.445 & 24.76 & 0.639 & 0.367 & 26.79 & 0.746 & 0.239  \\ 
zip-NeRF & 20.87 & 0.565 & 0.421 & 23.27 & 0.696 & 0.257 & \textbf{\textcolor{cyan}{29.66}} & 0.875 & \textbf{\textcolor{BurntOrange}{0.097}}  \\ 
3DGS & 20.71 & 0.619 & 0.394 & 23.05 & 0.740 & 0.243 & 29.19 & 0.880 & \textbf{\textcolor{cyan}{0.107}}  \\ 
3DGS+EWA & 23.70 & 0.667 & 0.369 & 25.90 & 0.775 & 0.236 & 29.30 & 0.880 & 0.114  \\ 
Mip-Splatting & \textbf{\textcolor{cyan}{26.47}} & \textbf{\textcolor{cyan}{0.754}} & \textbf{\textcolor{cyan}{0.305}} & \textbf{\textcolor{cyan}{27.39}} & \textbf{\textcolor{cyan}{0.808}} & \textbf{\textcolor{cyan}{0.205}} & 29.39 & \textbf{\textcolor{cyan}{0.884}} & 0.108  \\ 
Ours  & \textbf{\textcolor{BurntOrange}{28.42}} & \textbf{\textcolor{BurntOrange}{0.868}} & \textbf{\textcolor{BurntOrange}{0.238}} & \textbf{\textcolor{BurntOrange}{29.30}} & \textbf{\textcolor{BurntOrange}{0.870}} & \textbf{\textcolor{BurntOrange}{0.146}} & \textbf{\textcolor{BurntOrange}{29.84}} & \textbf{\textcolor{BurntOrange}{0.893}} & 0.112  \\ \hline
\end{tabular}
}
\vspace{-0.3cm}
\end{table*}

\FloatBarrier
\begin{figure}[ht]
  \centering
  \centering \includegraphics[width=1.0\textwidth,height=0.46\textheight]{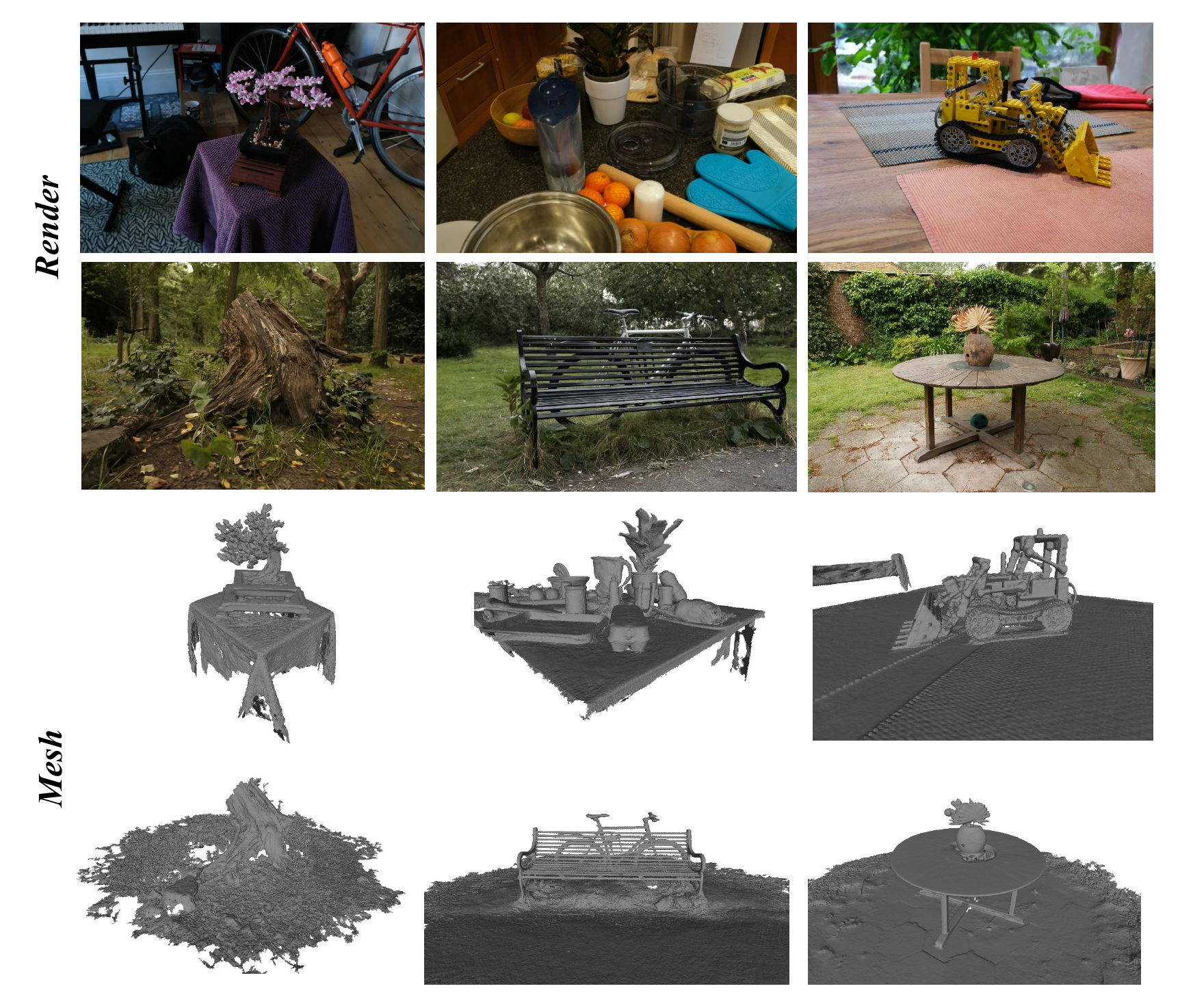}
   \caption{\textbf{Qualitative results on the Mip-NeRF360 dataset.} Our method reconstructs surfaces with fine geometry details and produces high-fidelity renderings on Mip-NeRF360 dataset. }
   \label{fig:a1}
\end{figure}

\section{Additional Qualitative Results}

\begin{figure}[ht]
  \centering
  \centering \includegraphics[width=1.0\textwidth,height=0.5\textheight]{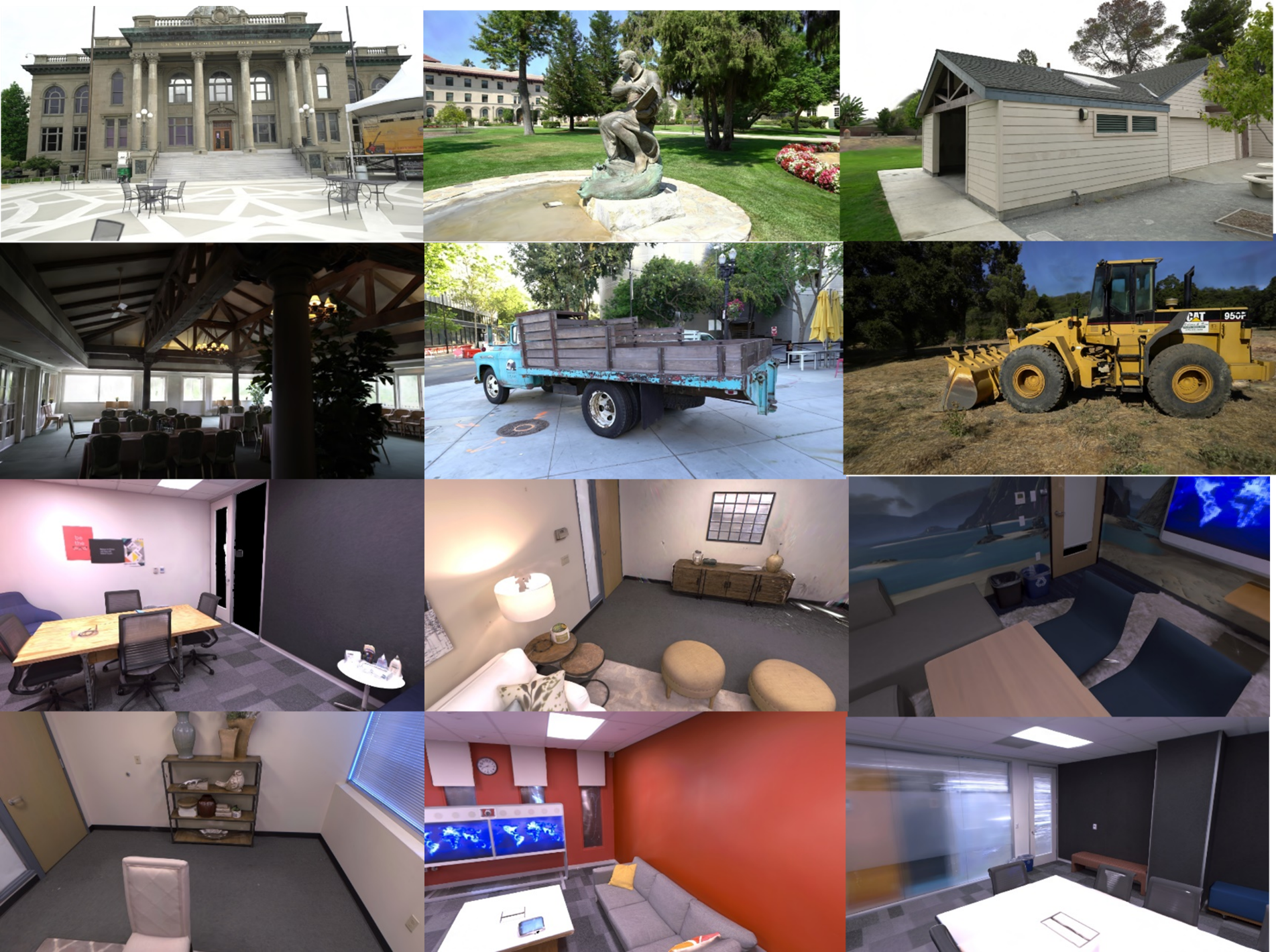}
   \caption{\textbf{Qualitative rendering results on TNT and Replica dataset.}  }
   \label{fig:a2}
\end{figure}

As shown in Fig.\ref{fig:1}, our method preserves fine-scale structures more accurately and achieves higher visual fidelity than 3DGS\cite{3dgs} and Mip-Splatting~\cite{mip360-splatting} across three representative scenes.
Fig.~\ref{fig:a1} displays the rendering (top) and surface reconstruction (bottom) results on the Mip-NeRF360~\cite{Mip-NeRF360} dataset. 
Additional rendering results on the TNT~\cite{TNT} and Replica~\cite{replica} datasets are provided in Fig.~\ref{fig:a2}. Collectively, these visual comparisons substantiate our method's capability for high-quality 3D reconstruction while maintaining critical geometric details.

\FloatBarrier

\section{Limitations}

Although our method can achieve high-resolution scene reconstruction under limited GPU resources, extending the block-based framework to dynamic scenes introduces new challenges. In particular, ensuring temporal consistency across frames and accurately modeling motion across spatial partitions remain open problems. 
Addressing these challenges will be a key focus of our future work, with the goal of enabling temporally coherent and spatially consistent reconstruction in dynamic environments.

\section{Broader Social Impacts}
\label{sec:impact}


This work presents Hierarchical Gaussian Splatting (HRGS), a super-resolution reconstruction framework tailored for computationally constrained environments. HRGS achieves high-fidelity 3D reconstruction while significantly reducing GPU memory usage and model size, thereby enhancing the accessibility of advanced 3D vision techniques. Its efficiency enables deployment on low-power platforms such as mobile and embedded devices, with promising applications in education, cultural heritage preservation, smart city visualization, and immersive virtual or augmented reality.
However, we acknowledge the potential for misuse in privacy-sensitive contexts, such as unauthorized spatial reconstruction. To mitigate such risks, the deployment of HRGS should be governed by clear ethical guidelines and regulatory oversight.


\newpage                                                                                                                                                                                                                                                                                                                                                             

\end{document}